%% file: main.tex
\documentclass[]{spie}
\usepackage[T1]{fontenc}
\usepackage{subcaption}
\usepackage{caption}
\usepackage{graphicx}
\usepackage{amssymb}
\usepackage{amsmath}
\usepackage{tikz}
\usepackage{wrapfig}
\usepackage{makecell}
\usepackage{array}
\usepackage{booktabs}
\usepackage{multirow}
\newcolumntype{C}[1]{>{\centering\arraybackslash}p{#1}}
\begin{document}

\title{Self-supervised Mamba-based Mastoidectomy Shape Prediction for Cochlear Implant Surgery}

\author[a]{Yike Zhang}
\author[a]{Eduardo Davalos}
\author[a]{Dingjie Su}
\author[b]{Ange Lou}
\author[a,b]{Jack H. Noble}
\affil[a]{Dept. of Computer Science, Vanderbilt University}
\affil[b]{Dept. of Electrical and Computer Engineering, Vanderbilt University}

% Option to view page numbers
\pagestyle{empty} % change to \pagestyle{plain} for page numbers   
\setcounter{page}{301} % Set start page numbering at e.g. 301
\maketitle

% abstract
\input{abstract}

% introduction
\input{intro}

% methodology
\input{method}

% experiment results
\input{results}

% conclusion
\input{conclusion}

% acknowledgement
\acknowledgments
This work was supported in part by grants R01DC014037 and R01DC008408 from the NIDCD. This work is solely the responsibility of the authors and does not necessarily reflect the views of this institute.

% ---- Bibliography ----
\bibliographystyle{spiebib}
\bibliography{citation}
\end{document}

%% file: abstract.tex
\begin{abstract}
% Cochlear Implant (CI) procedures involve inserting an array of electrodes into the cochlea located inside the inner ear. Mastoidectomy is a surgical procedure that uses a high-speed drill to remove part of the mastoid region of the temporal bone, providing safe access to the cochlea through the middle and inner ear. In this paper, we propose a mamba-based method to synthesize the mastoidectomy volume using only the preoperative Computerized Tomography (CT) scan, where the mastoid is intact. We introduce an self-supervised learning framework designed to predict the mastoidectomy shape. The goal of this paper is to reconstruct a 3D post-mastoidectomy surface directly from any given preoperative CT scan, aligning it with the intraoperative microscope view for various downstreaming tasks. Ear CT registration before surgery involves creating a mask to represent the mastoidectomy area and isolating the bone structure from the modified preoperative CT scans to generate a detailed mesh model. For model training purposes, this method uses postoperative CT scans to avoid manual data cleaning or labeling, even when the region removed during mastoidectomy is visible but affected by metal artifacts, low signal-to-noise ratio, or electrode wiring. Our approach estimates mastoidectomy regions with a mean Dice score of 0.70. 
% The code and models of our method are publicly available at:

Cochlear Implant (CI) procedures require the insertion of an electrode array into the cochlea within the inner ear. To achieve this, mastoidectomy, a surgical procedure involving the removal of part of the mastoid region of the temporal bone using a high-speed drill provides safe access to the cochlea through the middle and inner ear. In this paper, we propose a novel Mamba-based method to synthesize the mastoidectomy volume using only preoperative Computed Tomography (CT) scans, where the mastoid remains intact. Our approach introduces a self-supervised learning framework designed to predict the mastoidectomy shape and reconstruct a 3D post-mastoidectomy surface directly from preoperative CT scans. This reconstruction aligns with intraoperative microscope views, enabling various downstream surgical applications. For training, we leverage postoperative CT scans to bypass manual data cleaning and labeling, even when the region removed during mastoidectomy is affected by challenges such as metal artifacts, low signal-to-noise ratio, or electrode wiring. Our method achieves a mean Dice score of 0.70 in estimating mastoidectomy regions, demonstrating its effectiveness for accurate and efficient surgical preoperative planning.

\keywords{Self-supervised Learning, Mamba-based, Noisy Data, Mastoidectomy, Cochlear Implant, CT}
\end{abstract}

%% file: intro.tex
\section{Introduction}
Cochlear Implant (CI) procedures aim to restore hearing for patients with moderate-to-profound hearing disabilities \cite{labadie2018preliminary}. These surgeries require performing a mastoidectomy, a procedure that involves removing part of the temporal bone to access the cochlea. Dillon \cite{Dillon2015A} proposed an automatic method using a bone-attached robot to perform the mastoidectomy. A validation study was designed to show the robot can accurately perform the procedures, preserving critical anatomical structures. However, in their proposed pipeline, the surgeon must manually identify the mastoidectomy-removed region on a preoperative Computerized Tomography (CT) scan, which can be time-consuming and laborious. In this work, we aim to develop an automatic mastoidectomy prediction method. Such a method could be valuable for planning robotic surgeries or generating preoperative image-guided visualizations of the surgical field, supporting both surgical training and preoperative planning. We propose a deep learning-based segmentation network with a self-supervised training strategy. 
Using CT scans acquired after CI, the shape of the actual mastoidectomy can be visualized in a postoperative CT. Postoperative CT scans are often acquired with low-dose cone beam scanners, which sometimes have lower resolution and increased artifacts, making them more difficult to inspect visually. These scans suffer from metal artifacts (bright white noise caused by implant components) and intensity heterogeneity due to fluid accumulation in the ear canal, posing challenges for deep learning models without hand-drawn labels. 

Traditional approaches for predicting the mastoidectomy-removed region using preoperative/postoperative CT pairs for training, such as Generative Adversarial Networks (GANs)\cite{ciclegan} and diffusion-based models \cite{ddpm}, might not be suitable for our research goal due to the potential of generating undesired objects, such as fake electrodes and wires. Additionally, preoperative CTs have superior resolution compared to postoperative CTs in our dataset, delivering more detailed information for surgical planning and analysis. The removed mastoid bone presents an irregular and non-enclosed shape, as the mastoidectomy region is composed of pneumatized bone containing numerous small air cells. This structural complexity makes popular segmentation methods, such as SAM or SAM2-based models \cite{SAM4MIS, lou2024zeroshotsurgicaltoolsegmentation, lousam}, less suitable for this challenging task, given the ambiguous boundaries and irregular geometry of the mastoid region.
Therefore, a primary objective of our research is to design an effective framework to predict the mastoidectomy-removed region while preserving other original features of the preoperative CT and minimizing unnecessary alterations. Once the mastoidectomy shape has been successfully predicted, it can be applied to the original preoperative CT scans to generate a post-mastoidectomy CT mesh through isosurfacing, enabling accurate 3D reconstruction of the surgical outcome. The potential impact of building post-mastoidectomy CT surface is profound, as it provides a valuable resource for downstream tasks like surgical tool tracking, pose estimation of vital ear structures \cite{zhangmonocular}, and comprehensive surgical scene understanding and analysis \cite{lou2024surgicaldepthanythingdepth, zhang2024mastoidectomymultiviewsynthesissingle}.

The contributions of our work are highlighted as follows:
\begin{itemize}
    \item We are the first to use self-supervised learning technique to address the challenge of predicting mastoidectomy shapes directly from preoperative CT scans. By leveraging the power of self-supervised learning, our approach opens new possibilities for advancements in preoperative planning and mastoidectomy outcome prediction for the cochlear implant surgery.
    \item We build a comprehensive mastoidectomy shape prediction dataset from raw preoperative and postoperative medical images. To the best of our knowledge, this represents the first dataset specifically designed for this purpose, paving the way for advancing research in mastoidectomy shape prediction and related surgical applications.
    \item We propose a novel self-learning framework built on a cutting-edge Mamba-based architecture that outperforms other state-of-the-art models, including UNetr\cite{unetr} and SwinUNetr\cite{SwinUnetr}. Our method eliminates the need for manual annotation, offering an efficient and scalable training strategy.
\end{itemize}

%% file: method.tex
\section{Method}
\label{sec:method}
\begin{figure}[!ht]
  \centering
  \includegraphics[width=0.9\linewidth]{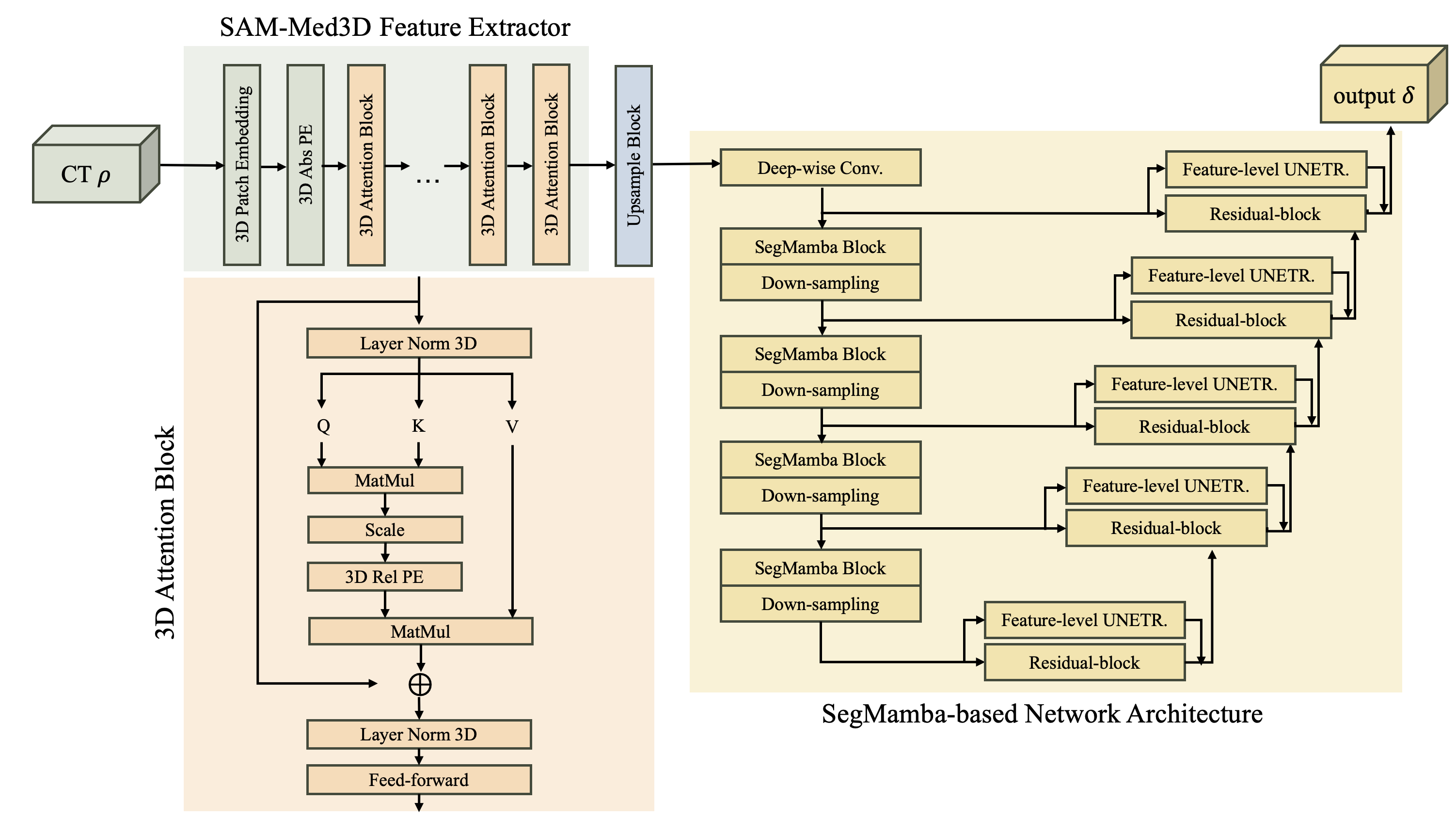}
  \caption{SegMamba-based Neural Network with pretrained SAM-Med3D Image Encoder}
  \label{fig:network_structure}
\end{figure}

In this work, we used a dataset of 751 preoperative and postoperative CT pairs from patients who underwent cochlear implantation. We randomly assigned 630 cases for training and validation, while reserving the rest for testing. We manually annotated the mastoidectomy-removed region in 32 randomly selected samples from the test dataset for evaluation. As preprocessing before training our method, the preoperative CT $\rho$ and postoperative CT $\omega$ were aligned via rigid registration, and their intensity values were normalized. 
As shown in Fig. \ref{fig:overview}, the removed mastoid volume in $\omega$ tends to have lower intensities compared to surrounding tissue since it only contains air and soft tissue. Thus, we develop a model to predict the mastoidectomy region (represented as an inverted probability map $\delta$) on $\rho$ and make the predicted post-mastoid CT ($\rho \otimes \delta$) have higher similarity when compared with $\omega$. The neural network shown in Fig. \ref{fig:network_structure} is modeled as a function $f_{\theta}$($\rho$) = $\delta$ using the state-of-the-art SegMamba-based \cite{xing2024segmambalongrangesequentialmodeling} neural network with a pretrained SAM-Med3D \cite{wang2023sammed3d} encoder as our feature extractor, where $\theta$ is the set of neural network parameters. In our model, we reduce the original CT dimensions to $[160, 160, 64]$ by cropping around the ear regions using an ear anatomy segmentation neural network proposed in Zhang et al.\cite{zhang2023} . 
\input{figures/overview}

The Multi-Scale Structural Similarity Index (MS-SSIM) \cite{msssim} represents an improvement over the traditional Structural Similarity Index (SSIM) \cite{ssim}. This enhancement enables MS-SSIM to comprehensively evaluate the similarities and differences between two images across multiple scales. It effectively accommodates the wide variety of heterogeneity that can occur between two different images. We extend this metric with Squared Cross-Correlation (SCC) \cite{scc} to quantify the overall structural similarity and distribution consistency between two image volumes $\rho \otimes \delta$ and $\omega$. We propose two self-supervised loss functions $L_{msssim\_cscc}$ and $L_{smooth}$ to train our model $f_{\theta}$. $L_{msssim\_cscc}$ effectively minimizes the impact of undesired noise present in postoperative CTs. However, this loss function may result in fragmentation within the output probability map, thus we add the second loss function $L_{smooth}$ to address the potential edge sharpness and encourage smoother value transitions in the output volume. The $L_{smooth}$ is defined as $L_{\text{smooth}}(\delta) = \sum_{i=1}^{N}\left\| \nabla \left( \delta_{i} \right) \right\|^2$. We define $\frac{\partial{\delta}}{\partial{x}} \approx \delta(i_x+1, i_y, i_z) - \delta(i_x, i_y, i_z)$, and the same rule applies to the $\frac{\partial{\delta}}{\partial{y}}$ and $\frac{\partial{\delta}}{\partial{z}}$. $L_{msssim\_cscc}$ performs across various scales, employing a multi-step down-sampling technique. Applying $L_{msssim\_cscc}$ alone effectively captures the target mastoidectomy region even when $\omega$ is contaminated by various noises. 
\input{equations/L_msssim_cscc}

%% file: figures/overview.tex
\begin{figure}[h]
\centering

\resizebox{0.9\textwidth}{!}{%
\tikzset{every picture/.style={line width=0.75pt}} %set default line width to 0.75pt        

\begin{tikzpicture}[x=0.75pt,y=0.75pt,yscale=-1,xscale=1]
%uncomment if require: \path (0,366); %set diagram left start at 0, and has height of 366

%Image [id:dp04985429909905015]
\draw (59.26,132.03) node  {\includegraphics[width=46.12pt,height=48.04pt]{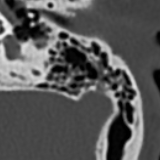}};
%Straight Lines [id:da7548488511529232] 
\draw    (90,140) -- (118,140) ;
\draw [shift={(120,140)}, rotate = 180] [fill={rgb, 255:red, 0; green, 0; blue, 0 }  ][line width=0.08]  [draw opacity=0] (12,-3) -- (0,0) -- (12,3) -- cycle    ;
%Shape: Rectangle [id:dp5075060272043457] 
\draw  [fill={rgb, 255:red, 248; green, 231; blue, 28 }  ,fill opacity=1 ] (120,96.79) -- (131.42,96.79) -- (131.42,170) -- (120,170) -- cycle ;
%Shape: Rectangle [id:dp9905265751013906] 
\draw  [fill={rgb, 255:red, 248; green, 231; blue, 28 }  ,fill opacity=1 ] (175.01,96.7) -- (186.43,96.7) -- (186.43,169.92) -- (175.01,169.92) -- cycle ;
%Straight Lines [id:da7318526237001484] 
\draw    (190,140) -- (218,140) ;
\draw [shift={(220,140)}, rotate = 180] [fill={rgb, 255:red, 0; green, 0; blue, 0 }  ][line width=0.08]  [draw opacity=0] (12,-3) -- (0,0) -- (12,3) -- cycle    ;
%Image [id:dp6245935950340661] 
\draw (248.82,132.03) node  {\includegraphics[width=46.78pt,height=48.04pt]{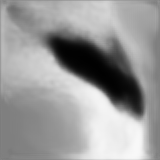}};
%Straight Lines [id:da4855113236477412] 
\draw    (285.07,140) -- (323.07,140) ;
\draw [shift={(325.07,140)}, rotate = 180] [fill={rgb, 255:red, 0; green, 0; blue, 0 }  ][line width=0.08]  [draw opacity=0] (12,-3) -- (0,0) -- (12,3) -- cycle    ;
%Flowchart: Summing Junction [id:dp7860787283221131] 
\draw   (330,140) .. controls (330,134.48) and (334.48,130) .. (340,130) .. controls (345.52,130) and (350,134.48) .. (350,140) .. controls (350,145.52) and (345.52,150) .. (340,150) .. controls (334.48,150) and (330,145.52) .. (330,140) -- cycle ; \draw   (332.93,132.93) -- (347.07,147.07) ; \draw   (347.07,132.93) -- (332.93,147.07) ;
%Straight Lines [id:da03850269106034809] 
\draw    (361.01,140) -- (388,140) ;
\draw [shift={(390,140)}, rotate = 180] [fill={rgb, 255:red, 0; green, 0; blue, 0 }  ][line width=0.08]  [draw opacity=0] (12,-3) -- (0,0) -- (12,3) -- cycle    ;
%Image [id:dp569968397908204] 
\draw (428.82,137.97) node  {\includegraphics[width=46.78pt,height=48.04pt]{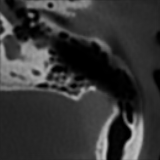}};
%Image [id:dp51617019234871] 
\draw (608.82,137.97) node  {\includegraphics[width=46.78pt,height=48.04pt]{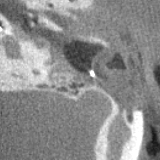}};
%Straight Lines [id:da508115717215246] 
\draw [color={rgb, 255:red, 65; green, 117; blue, 5 }  ,draw opacity=1 ] [dash pattern={on 4.5pt off 4.5pt}]  (467,140) -- (490,140) -- (490,140) -- (495,140) -- (568,140) ;
\draw [shift={(570,140)}, rotate = 180] [fill={rgb, 255:red, 65; green, 117; blue, 5 }  ,fill opacity=1 ][line width=0.08]  [draw opacity=0] (12,-3) -- (0,0) -- (12,3) -- cycle    ;
\draw [shift={(465,140)}, rotate = 0] [fill={rgb, 255:red, 65; green, 117; blue, 5 }  ,fill opacity=1 ][line width=0.08]  [draw opacity=0] (12,-3) -- (0,0) -- (12,3) -- cycle    ;
%Straight Lines [id:da5476263888087551] 
\draw [color={rgb, 255:red, 65; green, 117; blue, 5 }  ,draw opacity=1 ] [dash pattern={on 4.5pt off 4.5pt}]  (250.25,197.24) -- (250.02,172) ;
\draw [shift={(250,170)}, rotate = 89.48] [fill={rgb, 255:red, 65; green, 117; blue, 5 }  ,fill opacity=1 ][line width=0.08]  [draw opacity=0] (12,-3) -- (0,0) -- (12,3) -- cycle    ;
%Straight Lines [id:da3067217841729828] 
\draw    (60,80) -- (60,60) -- (340,60) -- (340,118) ;
\draw [shift={(340,120)}, rotate = 270] [fill={rgb, 255:red, 0; green, 0; blue, 0 }  ][line width=0.08]  [draw opacity=0] (12,-3) -- (0,0) -- (12,3) -- cycle    ;
%Shape: Rectangle [id:dp29554251195450276] 
\draw  [fill={rgb, 255:red, 248; green, 231; blue, 28 }  ,fill opacity=1 ] (136,113.18) -- (144.98,113.18) -- (144.98,159) -- (136,159) -- cycle ;
%Shape: Rectangle [id:dp18357304257344298] 
\draw  [fill={rgb, 255:red, 248; green, 231; blue, 28 }  ,fill opacity=1 ] (162,113.18) -- (170.98,113.18) -- (170.98,159) -- (162,159) -- cycle ;

% Text Node
\draw (146.96,124.12) node [anchor=north west][inner sep=0.75pt]   [align=left] {...};
% Text Node
\draw (134.21,72.4) node [anchor=north west][inner sep=0.75pt]    {$f_{\theta }( \rho )$};
% Text Node
\draw (241,82.4) node [anchor=north west][inner sep=0.75pt]    {$\delta $};
% Text Node
\draw (57,82.4) node [anchor=north west][inner sep=0.75pt]    {$\rho $};
% Text Node
\draw (411,82.4) node [anchor=north west][inner sep=0.75pt]    {$\rho \otimes \delta $};
% Text Node
\draw (601,81.4) node [anchor=north west][inner sep=0.75pt]    {$\omega $};
% Text Node
\draw (467,152.4) node [anchor=north west][inner sep=0.75pt]  [color={rgb, 255:red, 65; green, 117; blue, 5 }  ,opacity=1 ]  {$L_{sim}( \omega ,\ \rho \otimes \delta )$};
% Text Node
\draw (221,200.4) node [anchor=north west][inner sep=0.75pt]  [color={rgb, 255:red, 65; green, 117; blue, 5 }  ,opacity=1 ]  {$L_{smooth}( \delta )$};
\end{tikzpicture}}
\caption{Framework overview. We compare the predicted mastoidectomy-removed region applied to the preoperative CT $\rho \otimes \delta$ against the postoperative CT $\omega$ using similarity-based loss function $L_{msssim\_cscc}$. A smoothing term $L_{smooth}$ is also applied to $\delta$ to further reduce noise output by network $f_{\theta}$.}
\label{fig:overview} % Optional: if you want to reference the figure in your text
\end{figure}

%% file: equations/L_msssim_cscc.tex
% \begin{equation}
%     \textstyle \text{MSSSIM\_SCC}(\rho \otimes \delta, \omega) = [l_M(\rho \otimes \delta, \omega)]^{\alpha_M} \cdot \prod_{j=1}^{M} [c_j(\rho \otimes \delta, \omega)]^{\beta_j} [s_j(\rho \otimes \delta, \omega)]^{\gamma_j} SCC_j(\rho \otimes \delta, \omega)
% \label{Eq:L_msssim_scc}
% \end{equation} 
\begin{multline}
L_{\text{msssim\_cscc}}(\rho \otimes \delta, \omega) = 1 - [l_M(\rho \otimes \delta, \omega)]^{\alpha_M}
\cdot \prod_{j=1}^{M} [c_j(\rho \otimes \delta, \omega) + SCC_j(\rho \otimes \delta, \omega)]^{\beta_j} [s_j(\rho \otimes \delta, \omega)]^{\gamma_j}
\label{Eq:L_msssim_cscc}
\end{multline} In Eq. \ref{Eq:L_msssim_cscc}, $M$ represents the total number of scales at which the comparison is performed, and we assign $M$=5 in the loss function\cite{msssim}. $\alpha_M$ is the weight given to the luminance component at the coarsest scale. $\beta_j$ refers to the weights assigned to the contrast components at each scale $j$. $\gamma_j$ are the weights for the structure components at each scale $j$. l($\rho \otimes \delta$, $\omega$) is the luminance comparison function that measures the similarity in brightness between two image volumes. c($\rho \otimes \delta$, $\omega$) is the contrast comparison function that evaluates the similarity in contrast between two image volumes. s($\rho \otimes \delta$, $\omega$) is the structure comparison function that represents the similarity in structure or pattern between two image volumes. 
SCC is being added to the contrast comparison function in this equation to enhance MS-SSIM capability to further maximize the distribution similarity at each scaling level. SCC is defined as:
\begin{equation}
    \text{SCC}(
\rho \otimes \delta, \omega) = \frac{\Bigl(\sum_{i=1}^{N}\left [\left( (\rho \otimes \delta)_{i} - \overline{
\rho \otimes \delta} \right) \left( \omega_{i} - \overline{\omega} \right) \right ]\Bigl)^2}{\sum_{i=1}^{N}\left(  (\rho \otimes \delta)_{i} - \overline{
\rho \otimes \delta}  \right)^2 \sum_{i=1}^{N}\left(  \omega_{i} - \overline{\omega}  \right)^2},
\label{Eq:scc}
\end{equation} where $N$ is noted as the total number of pixels.

% Following the original work \cite{msssim}, we define l($\rho \otimes \delta$, $\omega$), c($\rho \otimes \delta$, $\omega$), and s($\rho \otimes \delta$, $\omega$) as:
% \begin{align}
% l(\rho \otimes \delta, \omega) &= \frac{2\mu_{\rho \otimes \delta}\mu_{\omega} + C_1}{\mu_{\rho \otimes \delta}^2 + \mu_{\omega}^2 + C_1}, \\
% c(\rho \otimes \delta, \omega) &= \frac{2\sigma_{\rho \otimes \delta}\sigma_{\omega} + C_2}{\sigma_{\rho \otimes \delta}^2 + \sigma_{\omega}^2 + C_2}, \\
% s(\rho \otimes \delta, \omega) &= \frac{\sigma_{\rho \otimes \delta \otimes \omega} + C_3}{\sigma_{\rho \otimes \delta}\sigma_{\omega} + C_3}.
% \end{align} Here, $\mu_{\rho \otimes \delta}$ and $\mu_{\omega}$ are the mean luminance values of image volumes $\rho \otimes \delta$ and $\omega$. $\sigma_{\rho \otimes \delta}$ and $\sigma_\omega$ represent the standard deviations of the two image volumes, respectively. $\sigma_{\rho \otimes \delta \otimes \omega}$ is the covariance between the two image volumes that show structural similarity. $C_1$, $C_2$, and $C_3$ are small constants added to the equation to avoid division by zero. 

%% file: results.tex
\section{Results}
\begin{figure}[h!]
  \centering
  \includegraphics[width=0.9\linewidth]{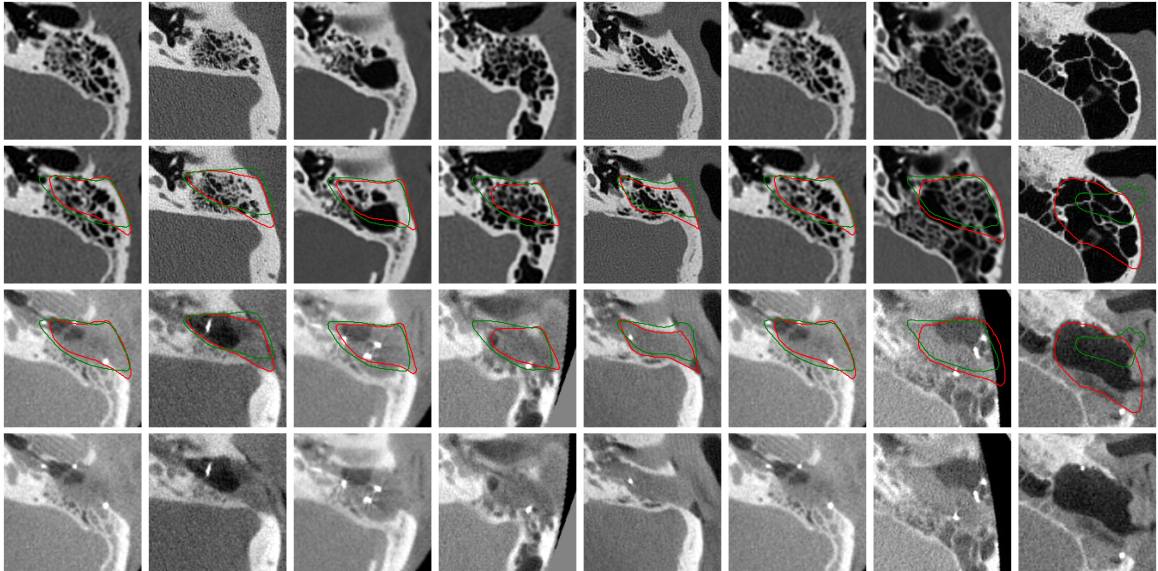}
\caption{Qualitative Performance Evaluation. The first row displays preoperative CT scans. The second row overlays predicted mastoidectomy-removed areas (green) and ground truth (red) on preoperative CTs. The third row shows these contours on postoperative CT scans, and the final row displays the postoperative CTs. }
\label{fig:qualitative}
\end{figure}
Our dataset comprises a total of 751 images, divided into 504 training samples, 126 validation samples, and 121 testing samples. As stated in Section \ref{sec:method}, we manually annotated 32 ground truth mastoidectomy volume labels from the testing dataset. This annotation process, requiring approximately one hour per case, was highly labor-intensive due to the inherent heterogeneity between preoperative and postoperative CT scans, as well as the substantial variability in the shape and size of the removed mastoid regions. To assess the proposed model's performance in predicting mastoidectomy shapes within preoperative CT scans, we input voxel values from the 3D preoperative images into the Mamba-based model and binarize the resulting probability masks to generate the final predictions. Fig. \ref{fig:qualitative} shows the qualitative performance on eight test samples, with each column representing a test case. Our method predicts the mastoidectomy-removed region well in the majority of the cases. The last column shows the case with the worst Dice score using our proposed method. In this case, the mastoid region in that preoperative CT scan is unusually sparse compared to the other samples in our test dataset. This leads to a predicted mastoidectomy region that is smaller than the ground truth using the self-supervised method.

\begin{figure}[h!]
  \centering
  \includegraphics[width=\linewidth]{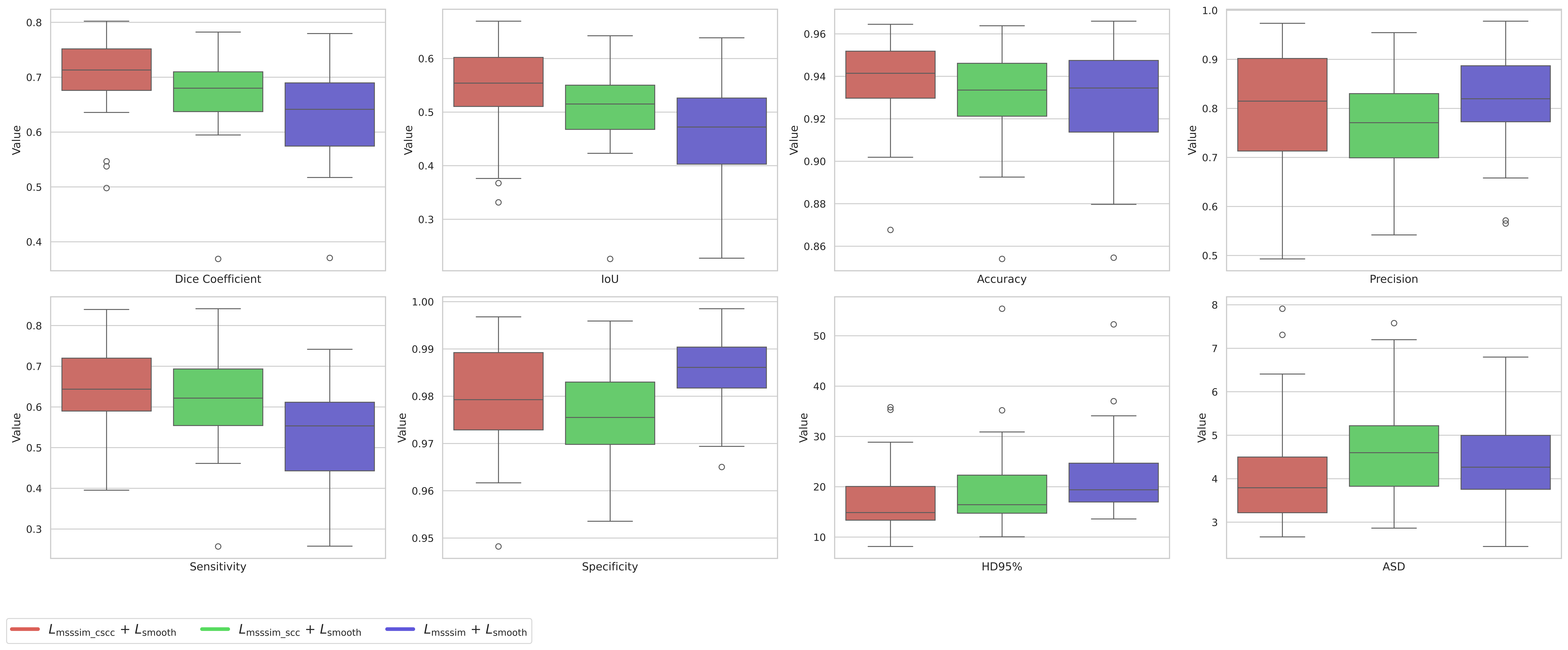}
\caption{Ablation Study: comparing performance across different loss functions.}
\label{fig:ablation_study}
\end{figure}

Additionally, Sørensen–Dice coefficient (Dice), Intersection over Union (IoU), Accuracy (Acc), Precision (Pre), Sensitivity (Sen), Specificity (Spe), Hausdorff Distance (HD95\%), and Average Surface Distance (ASD) are used to measure the similarities between the predicted mastoidectomy-removed region by our neural network and ground truth labels in Table \ref{Tab:analysis}. The HD95\% metric is a modification of the Hausdorff Distance that focuses on the 95th percentile of the distance to reduce the influence of outliers. As shown in the table, it compares the aforementioned metrics of widely-used transformer-based networks and U-Net-based networks, including SwinUNetr \cite{SwinUnetr}, UNETR \cite{unetr}, the U-Net++ \cite{zhou2019unetplusplus} / U-Net \cite{unet} network using ImageNet \cite{imagenet} pretrained weights, and our proposed mamba-based method. The asterisk $^*$ followed by a number indicates that the difference between the proposed and competing methods is statistically significant based on a Wilcoxon test with a \textit{p}-value of less than 0.05. We observe that the predictions obtained by the proposed method tend to have lower sensitivity than specificity and precision. This suggests that our model is more likely to produce false negative values than false positive values, implying that the synthesized regions are generally smaller than the ground truth labels. Fig. \ref{fig:ablation_study} shows our method's ablation study with different loss functions. $L_{msssim\_cscc}$ is our proposed loss term, while $L_{msssim\_scc}$ refers to adding Eq. \ref{Eq:scc} to the s($\rho \otimes \delta$, $\omega$) structural similarity function instead of the c($\rho \otimes \delta$, $\omega$) contrast comparison function in MS-SSIM (Eq. \ref{Eq:L_msssim_cscc}). $L_{msssim}$ indicates the original MS-SSIM loss function. As shown in the figure, the proposed loss term outperforms the other combinations across the majority of metrics. 
\input{table/metric}

\begin{figure}[h]
    \centering
    \begin{subfigure}[b]{0.4\textwidth}
        \centering
        \includegraphics[width=0.7\textwidth]{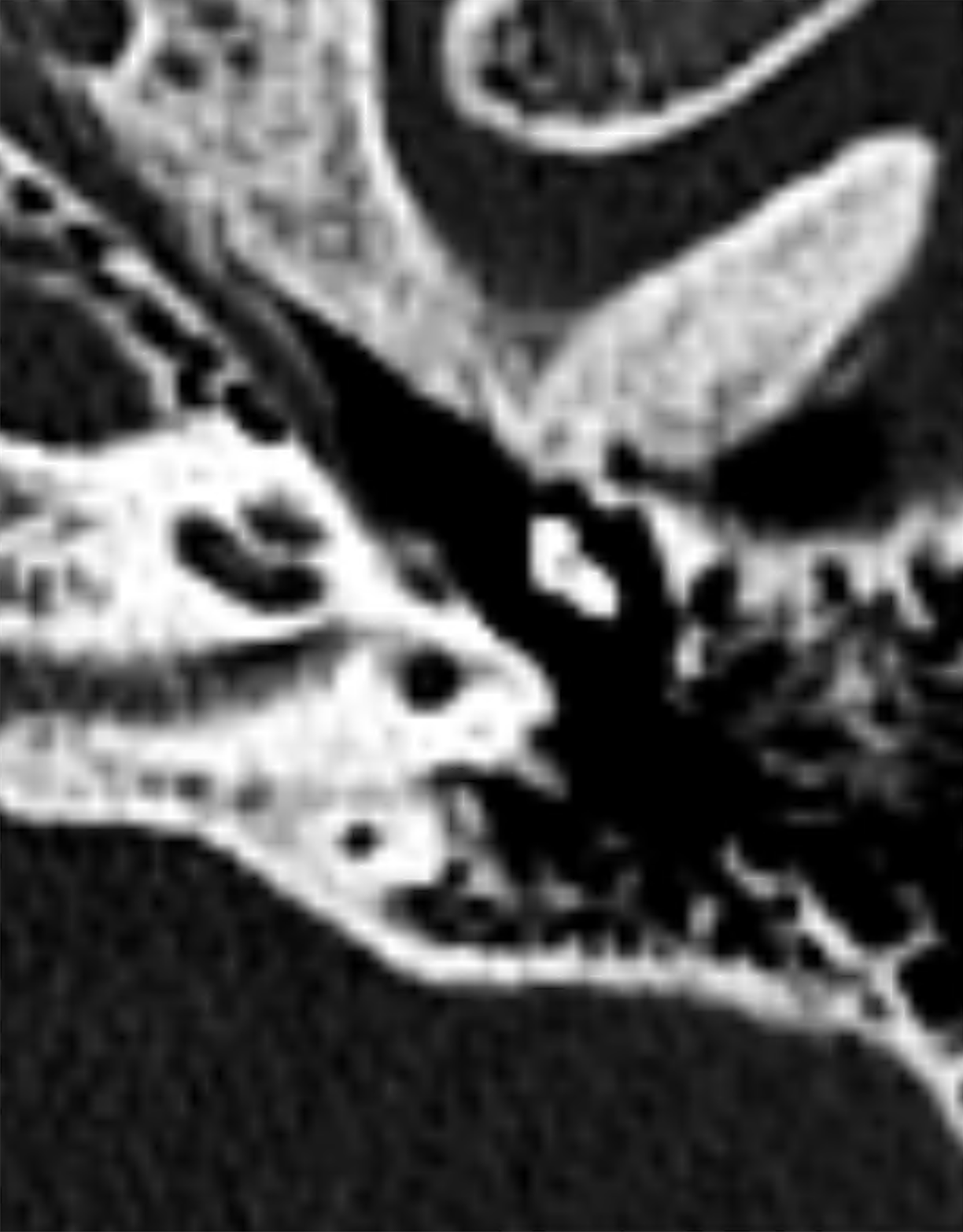}
        \caption{Original preoperative CT scan}
        \label{fig:ct}
    \end{subfigure}
    \begin{subfigure}[b]{0.4\textwidth}
        \centering
        \includegraphics[width=0.7\textwidth]{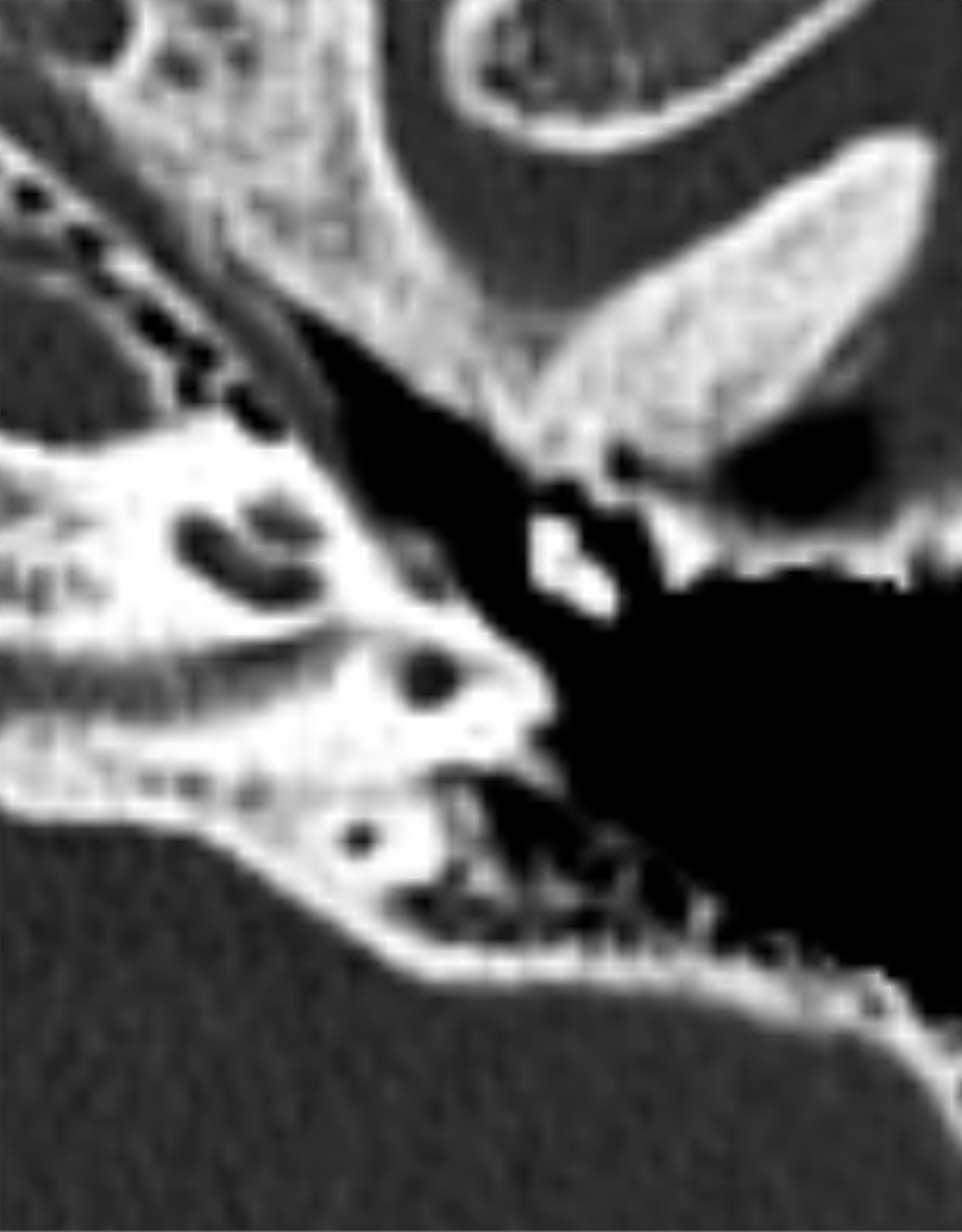}
        \caption{Modified preoperative CT scan}
        \label{fig:modified_ct}
    \end{subfigure}
    \begin{subfigure}[b]{0.4\textwidth}
        \centering
        \includegraphics[width=0.7\textwidth]{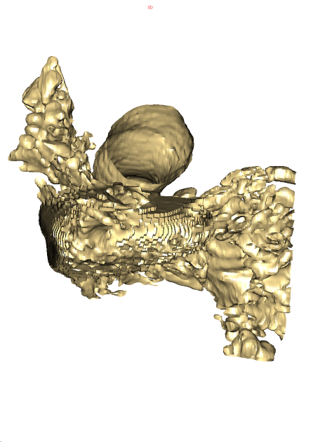}
        \caption{Reconstructed surface}
        \label{fig:ct_mesh}
    \end{subfigure}
    \begin{subfigure}[b]{0.4\textwidth}
        \centering
        \includegraphics[width=0.7\textwidth]{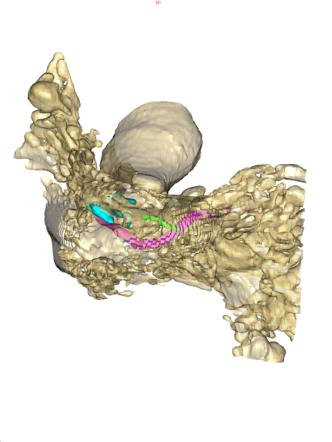}
        \caption{Ear structures overlay}
        \label{fig:meshes_overlay}
    \end{subfigure}
    \caption{Reconstructed post-mastoidectomy surface from the modified preoperative CT scan with the predicted mastoidectomy region.}
    \label{fig:extracted_meshes}
\end{figure}
After the successful prediction of the mastoidectomy shape region using our proposed self-supervised learning method, we provide an additional example of post-mastoidectomy CT surface reconstruction in Fig. \ref{fig:extracted_meshes}, achieved by applying isosurfacing to the preoperative CT scan modified with the predicted mastoidectomy shape. Specifically, Fig. \ref{fig:ct} shows the original preoperative CT scan. Fig. \ref{fig:modified_ct} illustrates the preoperative CT scan masked with the predicted mastoidectomy region. Fig. \ref{fig:ct_mesh} displays the corresponding CT mesh reconstructed using the isosurface function applied to the modified preoperative CT scan. In Fig. \ref{fig:meshes_overlay}, the reconstructed surface is shown with an overlay of essential ear structures, including the ossicles, facial nerve, and chorda, providing a comprehensive visualization of the overall post-mastoidectomy CT reconstruction results.

%% file: table/metric.tex
\begin{table*}[ht]
    % Table formatting
    \centering
    \footnotesize
    \begin{center}
        \begin{tabular}{ |l|c|c|c|c|c|c|c|c| }
            \hline
            \multicolumn{1}{|c|}{Methods} &
            \multicolumn{1}{c|}{Dice $\uparrow$} & 
            \multicolumn{1}{c|}{IoU $\uparrow$} & 
            \multicolumn{1}{c|}{Acc $\uparrow$} & 
            \multicolumn{1}{c|}{Pre $\uparrow$} & 
            \multicolumn{1}{c|}{Sen $\uparrow$} &
            \multicolumn{1}{c|}{Spe $\uparrow$} & 
            \multicolumn{1}{c|}{HD95\% $\downarrow$} &
            \multicolumn{1}{c|}{ASD $\downarrow$} \\
            \hline
            UNet\cite{unet} & 0.5432$^{*}$ & 0.3867 & 0.9203 & 0.8215 & 0.4319 & 0.9877 & 26.1439 & 5.0094$^{*}$\\
            \hline
            UNet++\cite{zhou2019unetplusplus} & 0.6527$^{*}$ & 0.4931 & 0.9230 & 0.7937 & 0.6010 & 0.9671 & 20.3570 & 6.2326$^{*}$\\
            \hline
            UNetr\cite{unetr} & 0.5608$^{*}$ & 0.4020 & 0.9263 & \textbf{0.8734} & 0.4246 & \textbf{0.9936} & 20.1116 & 6.5478$^{*}$\\
            \hline
            SwinUNetr\cite{SwinUnetr} & 0.6426$^{*}$ & 0.4834 & 0.9317 & 0.8506 & 0.5293 & 0.9885 & \textbf{17.2194} & 5.6297$^{*}$\\
            \hline
            Our Method  & \textbf{0.7019} & \textbf{0.5450} & \textbf{0.9375} & 0.7970 & \textbf{0.6444} & 0.9795 & 17.3515 & \textbf{4.1585}\\
            \hline
        \end{tabular}
    \end{center}
    % Table's information
    \caption{Performance comparison between Transformer/U-Net-based network and Mamba-based network}
    \label{Tab:analysis}
\end{table*}

%% file: conclusion.tex
\section{Discussion and Conclusion}
Our paper introduces a novel self-supervised mamba-based deep learning framework and a new loss function that combines MSSSIM and SCC for mastoidectomy in CI surgery using highly noisy original postoperative CT datasets. We propose an innovative approach to train neural networks with only noisy data, eliminating the need for extensive manual cleaning and labeling. This methodology can potentially extend to other surgical procedures where manual annotations are challenging and time-consuming. The proposed self-supervised learning framework predicted the mastoidectomy region with a mean Dice coefficient of 0.70. These results show promise that our self-supervised network can assist surgeons in planning a mastoidectomy without requiring extensive and laborious manual annotation. As discussed in this paper, reconstructing the post-mastoidectomy surface enables a range of downstream tasks, such as developing intraoperative navigation technology for cochlear implant surgery, advancing 3D surgical scene understanding, and providing detailed surgical analysis. In the future work, we aim to enhance the plain post-mastoidectomy surface by adding realistic textures to it, further improving its resemblance to actual surgical views and generating novel synthetic multi-views for cochlear implant surgery.